\documentclass[10pt,twocolumn,letterpaper]{article}

\usepackage{cvpr}
\usepackage{times}
\usepackage{epsfig}
\usepackage{graphicx}
\usepackage{amsmath}
\usepackage{amssymb}
\usepackage{commath}
\usepackage{caption} 
\usepackage{subcaption}
\usepackage{appendix}

\newenvironment{packed_enum}{
\begin{enumerate}
  \setlength{\itemsep}{1pt}
  \setlength{\parskip}{0pt}
  \setlength{\parsep}{0pt}
}{\end{enumerate}}

\usepackage[breaklinks=true,bookmarks=false]{hyperref}

\cvprfinalcopy 

\def\cvprPaperID{****} 

\begin{document}

\title{EyeNet: A Multi-Task Network for Off-Axis Eye Gaze Estimation and \\ User Understanding}


\author{Zhengyang Wu\quad Srivignesh Rajendran\quad Tarrence van As \quad Joelle Zimmermann \\ Vijay Badrinarayanan \quad Andrew Rabinovich\\ Magic Leap, Inc. \\ \tt\small \{zwu,srajendran,tvanas,jzimmermann,vbadrinarayanan,arabinovich\}@magicleap.com} 

\maketitle

\begin{abstract}
Eye gaze estimation and simultaneous semantic understanding of a user through eye images is a crucial component in Virtual and Mixed Reality; enabling energy efficient rendering, multi-focal displays and effective interaction with 3D content. In head-mounted VR/MR devices the eyes are imaged off-axis to avoid blocking the user's gaze, this view-point makes drawing eye related inferences very challenging. In this work, we present EyeNet, the first single deep neural network which solves multiple heterogeneous tasks related to eye gaze estimation and semantic user understanding for an off-axis camera setting. The tasks include eye segmentation, blink detection, emotive expression classification, IR LED glints detection, pupil and cornea center estimation. To train EyeNet end-to-end we employ both hand labelled supervision and model based supervision. We benchmark all tasks on MagicEyes, a large and new dataset of $587$ subjects with varying morphology, gender, skin-color, make-up and imaging conditions.
\end{abstract}

\section{Introduction}
Eye gaze estimation and simultaneous understanding of the user, through eye images, is a critical component for current and future generations of head-mounted devices (HMDs) for virtual and mixed reality.  It enables energy and bandwidth efficient rendering of content (foveated rendering~\cite{guenter2012foveated}), drives multi-focal displays for more realistic rendering of content (minimizing accommodation vergence conflict~\cite{oculustalk}), and provides an effective and non-obtrusive method for understanding user's expressions.

\begin{figure}[h!]
  \includegraphics[width=0.5\textwidth,height=\textheight,keepaspectratio]{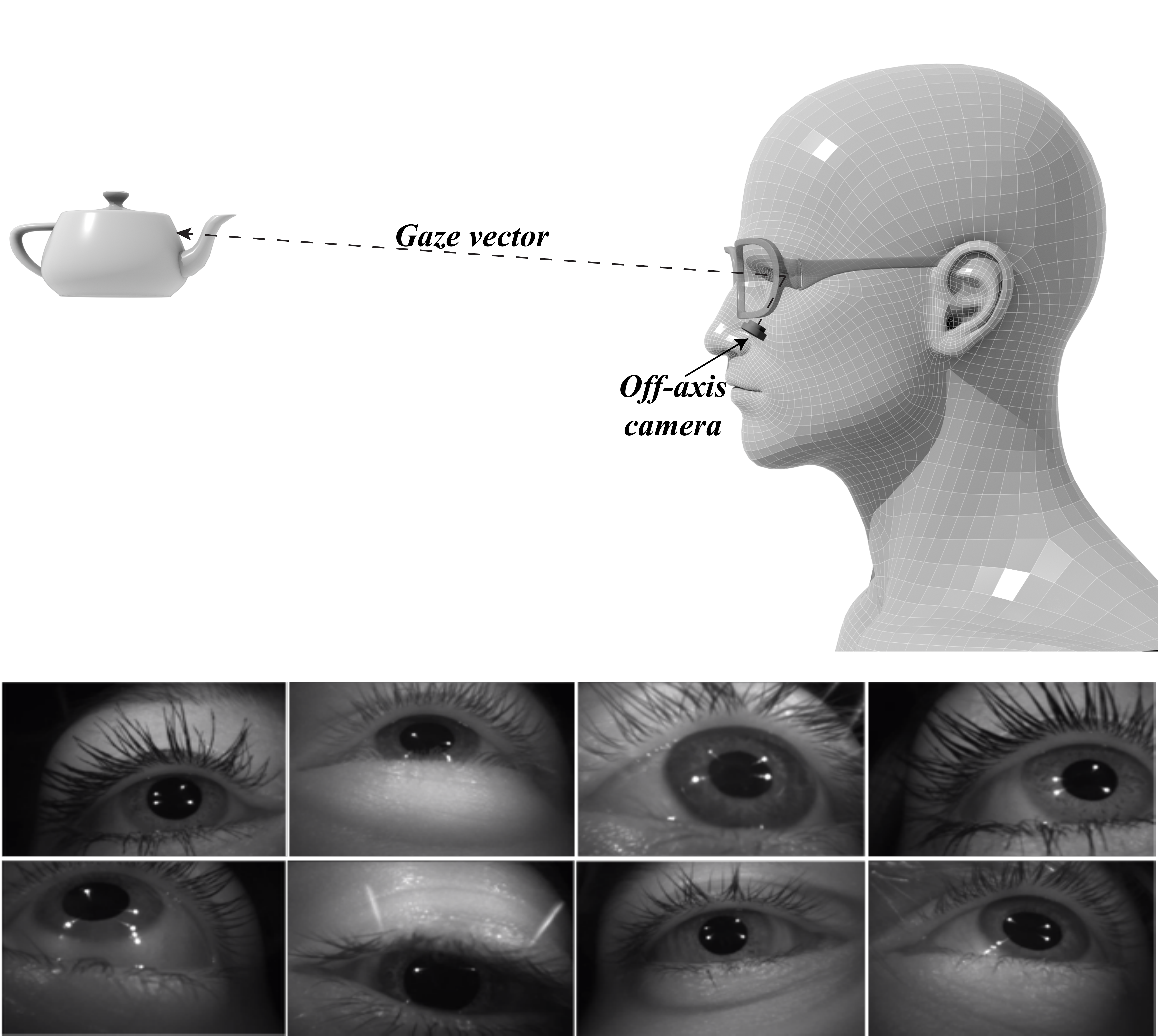}
  \caption{Off-axis IR camera setting in head-mounted VR/MR devices. Also shown are some sample eye images captured in this setting. }
  \label{fig:device}
\end{figure}


The typical setup in head-mounted Virtual and Mixed Reality devices for eye gaze estimation is illustrated in Figure~\ref{fig:device}. A set of four IR LEDs are placed in and around the display and their reflections (glints) are detected using the IR-sensitive eye camera, one for each eye. These glints are used to estimate important geometric quantities in the eye which are not directly observable from the eye camera images. A set of example images from such a setup is also shown in Figure~\ref{fig:device}. The four main semantic classes, Face/Background, Sclera, Iris and Pupil, along with the glints are clearly visible. As can be seen from these examples, there can be a large angle between the user's gaze and the camera axis. This makes eye gaze estimation challenging due to the increased eccentricity of pupils, partial occlusions caused by the eyelids and eyelashes~\cite{swirski2012robust}, as well as glint distractions caused due to environment illumination. 

\begin{figure}[h!]
  \includegraphics[width=0.48\textwidth,height=\textheight,keepaspectratio]{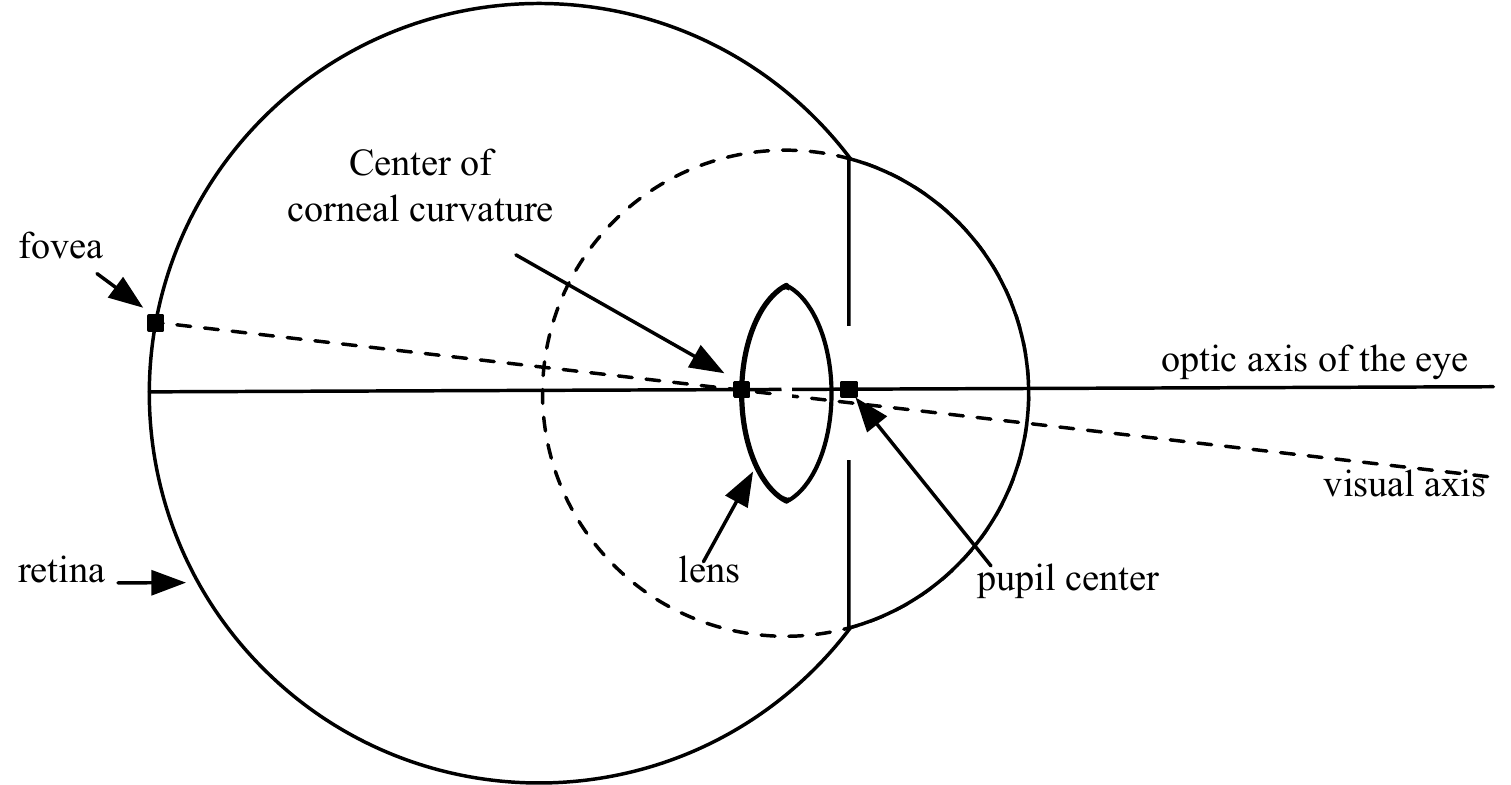}
  \caption{An illustration of the standard geometric model of the physiology of human eye.}
  \label{fig:Setup}
\end{figure}

A standard geometric model of the human eye is shown in Figure~\ref{fig:Setup}. The eye ball sphere encompasses the inner corneal sphere and within the corneal sphere lies the pupil opening. The \textit{optical axis} of the eye connects the cornea center and the pupil (opening) center. The \textit{visual axis} or \textit{gaze} vector, for all practical purposes, is taken to be the line joining the cornea center and the fovea at the back of the eye. The angle ($\kappa$) between the gaze vector and optical axis is assumed to be constant for each user. Estimating the optical axis (pupil and cornea center) is the key problem underlying gaze tracking. Per subject calibration can be performed to transform the optical axis to the gaze vector.

\begin{figure*}[h!]
  \includegraphics[width=\textwidth,height=\textheight,keepaspectratio]{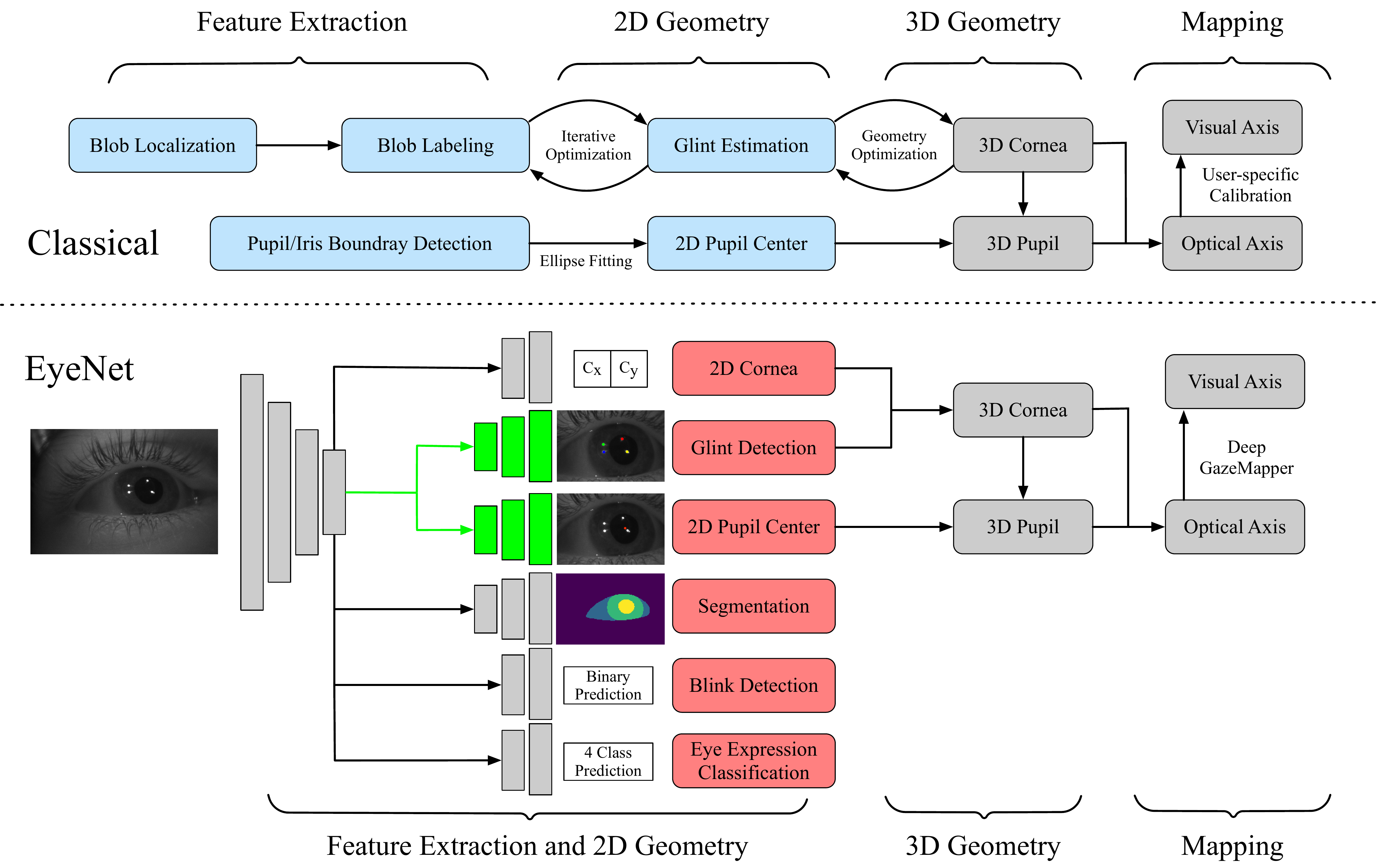}
  \caption{Eye Tracking Pipeline. The top part illustrates general stages of eye tracking pipelines. Middle part shows the stages of the ubiquitous classical baseline. Blue boxes indicate the classical counterparts that are replaced by EyeNet. Bottom part shows a simplified (for illustration purpose) version of our network architecture, where green boxes indicate weight sharing. Red boxes are the outputs of EyeNet with a single feature encoder network and six task decoders. Grey boxes corresponding to 3D geometry and gaze mapping in both pipelines share the same geometric model of the eye but differ in ways they utilize the model to estimate the desired quantities. (This figure is best viewed in color).}
  \label{fig:pipeline}
\end{figure*}

Accurate gaze (optical axis) estimation involves appearance based computation (detection of 2D attributes such as pupil and glints on the image) followed by geometry based computation of cornea center, pupil center, and gaze vector in 3D. Figure~\ref{fig:pipeline} illustrates the different stages for both classical (ubiquitous) and learned (EyeNet) eye tracking pipelines. 

The classical -- geometry-based -- eye trackers rely on hand engineered computer vision techniques to detect white blobs for localizing glints~\cite{guestrin2006general} and boundaries for pupil and iris~\cite{swirski2012robust} with many manually tuned thresholds. Following the appearance based estimation, in the geometric stage, the process of going from 2D image features to 3D gaze involves an iterative optimization module which detects glints and estimates the corneal center alternately. Given two or more LED locations and their corresponding glint pairs, the iterative optimization will estimate the 3D cornea center. Finally, to compute the optical axis, the detected 2D pupil center is back projected to 3D given the estimated 3D cornea center and assumptions about the eye geometry. Here we refer the reader to~\cite{guestrin2006general} for the complete derivations of the eye geometry based cornea, pupil estimation methods. 

EyeNet as compared to the classical pipeline is a multi-tasking deep neural network. We show that by training such a network appropriately it is possible to entirely eliminate the need to hand design heuristic methods to detect image features. Furthermore, EyeNet can simultaneously estimate robust initial values for glint positions and cornea center, after which a standard gradient descent optimization results in accurate gaze estimation. The shared learnt representation across several tasks enables robust estimation of all the desired quantities for gaze estimation while also amortizing the computational load. 

Semantic user understanding from eye images involve tasks such as accurate blink detection to enable multi-focal displays to switch from one focal depth to another to reduce accomodation-vergence mismatch~\cite{oculustalk}, emotive expression classification for fine-grained avatar animation for telepresence and to enable more nuanced communication in mixed reality. Semantic segmentation of the eyes by itself is useful to animate the eyes and eyebrows of avatars.

In this work, we demonstrate that it is indeed possible to share computation across both appearance based tasks (segmentation, glint, blink and emotive expression detection) and 2D geometry computation (cornea, pupil) while delivering accurate semantics and robust gaze estimates as compared to stand alone geometric gaze estimation methods. This work is also contemporaneous with other attempts in computer vision to learn multi-tasking networks for scene understanding~\cite{zamir2018taskonomy}. 


To summarize, our contributions in this paper are as follows:
\begin{packed_enum}
\item EyeNet: The first multi-task deep neural network trained to jointly estimate multiple quantities relating to eye gaze estimation and semantic user understanding from off-axis eye images.
\item A computationally efficient and robust estimation approach arising due to shared feature representation across both appearance based and geometric tasks which is useful for resource constrained applications.
\item Benchmarking the performance of EyeNet on a large and diverse dataset including variations in gender, race and physiology.
\item The first eye tracking dataset with all intermediate ground truth for a diverse demographic of gender, race and physiology.
\end{packed_enum}
\section{Related Work}
The problem of eye gaze or point of regard (PoR) estimation~\cite{guestrin2006general} is most commonly studied in the context of two application scenarios. The first is for monitoring user attention or saliency for content that is projected in modern electronic devices such as laptops or phones~\cite{krafka2016eye,kimconvolutional,huang2018predicting}. These devices are held at a fixed (often known) distance from the user, and are able to image the whole face with good resolution. The task then is narrowed down to estimating which part of the screen (x-y coordinates) the gaze is directed towards. Recent work has shown successful application of CNNs for this task~\cite{krafka2016eye,kimconvolutional}. ~\cite{krafka2016eye} in particular proposes a CNN trained on a large dataset of 2.5M frames collected from $\sim$1500 subjects, and demonstrates a tracking error of $1-2cm$. In general, CNNs are being successfully employed for saliency detection in images~\cite{kruthiventi2015deepfix,CAT2000,Judd_2012,huang2018predicting}. The common theme behind all these approaches is that the estimate is a 2D position or saliency map, and all of these rely on \textit{appearance-only} eye gaze estimation.

A second and different line of approach to eye gaze estimation is to infer geometric quantities of the eye such as pupil and cornea centers, and use them to estimate the optical/visual axis (gaze) in 3D~\cite{guestrin2006general}. These methods typically rely on appearance based features (pupil and iris boundaries, glints, etc) to estimate geometric quantities, based on assumptions from a geometric eye model~\cite{villanueva2008geometry}. The overall aim of these eye tracking systems is to estimate the 3D gaze vector and the vergence/fixation/point of regard in 3D without recourse to a screen. Another important difference with respect to the previous class of methods is that such systems~\cite{tobiieyetracker} do not image the entire face, and instead have separate cameras to image each eye (similar to the setup in Figure~\ref{fig:device}). Having individual low resolution eye cameras that are off the visual axis makes the problem much more challenging (increased eccentricity of the pupil/iris boundaries, occlusions due to eye-lids and eye-lashes). For such approaches with an off-axis camera setup data driven end-to-end solutions seem to suffer from low precision~\cite{shrivastava2017learning}.

Multiple approaches have been proposed to overcome the challenges posed by the off-axis eye tracking setup. Recent analysis-by-synthesis approach~\cite{wood2014eyetab} uses large-scale procedurally rendered synthetic data for nearest neighbor based gaze estimation. While the results look encouraging, the precision is not ideal. Shape based methods~\cite{tong2010unified,hansen2005eye} use pupil and iris boundary shape to compute gaze. The use of traditional computer vision techniques to estimate region boundaries makes them fragile, with failures frequently occurring under lighting variations and low image quality.  The Corneal reflection methods use external light sources (LED) to create a pattern of glints on each eye~\cite{zhu2007novel,guestrin2006general,villanueva2008geometry,yoo2005novel,hennessey2006single}. The location of the LEDs and their corresponding glint reflections in the eye are used to estimate the position of cornea in 3D. This along with pupil position estimation provides the optical axis of the camera. These methods tend to have the highest precision in gaze estimation but rely on traditional image processing to estimate pupil position and glint locations, which compromises their robustness. 

A recent work aims to train deep neural networks for low latency gaze estimation using largely synthetically generated and rendered eye images~\cite{kim2019nvgaze}. The results show the promise of using synthetic data at large scale to perform end to end learning when close attention is paid to eye modeling and rendering images according to real camera characteristics. EyeNet itself could benefit from training on such synthetic data when available. However, from the point of view of gaze estimation, their experiments were restricted to on-axis images with real data collected in controlled settings which reduce noise in the ground truth. In addition, their best reported results use per subject training and testing on very few subjects. In practice though the data collected in the wild for any user is usually noisy. Unlike the data displayed on a screen at a fixed distance, as done in their experiments, real calibration targets are virtual objects in complex environments. When users are prompted to focus on them, invariably there exist errors due to loss of concentration and distraction, and hence the quality of the ground truth gaze is noisy. Overall, in comparison, our experiments were conducted on a much larger scale and in realistic settings representative of those found in the wild. Even when noisy ground truth is used for training, our results are competitive with state of the art trackers while simultaneously providing a variety of useful inferences.

In this work, we aim to exploit the robustness of the appearance based approaches and the precision of the geometric methods (corneal reflection) to contribute to the final gaze estimation. To this end, our proposed EyeNet is trained using supervision from human labelers to estimate boundaries (segmentation), temporal events (blinks), key points (pupil, glint detection). 3D cornea position training uses the geometric eye model based supervision. The use of a single network with a shared feature representation across these correlated tasks (for a particular gaze, boundaries, glints,  pupil position, cornea position are all related) provides implicit regularization leading to robustness.


\section{Classical Eye Gaze Estimation}
The algorithmic pipeline used in the classical gaze estimation pipeline can be seen in Figure~\ref{fig:pipeline}. In our implementation of this pipeline the first step is to detect the 
pupil, iris and glint boundaries in the input eye images. To obtain the pupil and iris boundaries we use a trained deep eye parts segmentation network (see Section~\ref{sec:eye_parts_segmentation}) which provides robust and accurate pixel-wise segmentation. A pupil ellipse is fit to the pupil segment boundary to derive the 2D pupil center. The iris boundary is used to restrict search for bright blobs corresponding to LED glints. Binary maps are created by adaptive thresholding using the image histogram. Then basic image processing steps are used~\cite{OpenCVBlob} to obtain bright blobs. Note that although there are only 4 LEDs on the HMD used in our experiments the number of detected blobs can be greater or lesser than 4. An ellipse is then fit to each blob to obtain their centers.
In this pipeline, the 3D position of the cornea center and detection of the glints (location and association to the correct LED i.e labelling) given the blob centers is done simultaneously using iterative optimization based on standard back-projection losses (see Section~\ref{sec:geometry} for the details). The estimate of the 3D cornea center is then used to lift the 2D pupil center to 3D, once again by back projecting the 2D pupil center estimate to intersect with the cornea ball of fixed radius. 
The final step is to use 3D cornea and pupil estimates (i.e the optical axis) and fit a standard polynomial mapper learned from calibration frames to map the optical axis to the gaze direction~\cite{villanueva2008geometry}.
 Note that in this pipeline there are several hand engineered heuristic modules to detect glints, pupil center, cornea center and reject gaze estimation etc. These heuristics need to be updated by experts when new data arrives. In contrast, the trained EyeNet model can robustly predict the required geometric quantities in a feed forward manner using convolutions. Furthermore, the model weights can be updated as new data arrives by finetuning. This simplifies both implementation and code maintenance.
 
\section{Eye Net}

Gaze estimation has been shown to be a challenging learning task for deep networks. Recent end-to-end learning approaches~\cite{krafka2016eye, shrivastava2017learning} have achieved relatively low gaze tracking precision. While recent work~\cite{kim2019nvgaze} has sought to overcome this with careful synthetic eye renderings it has mainly done so for an on-axis camera setting and relatively noise free real data. The low precision for end-to-end approaches in real world settings can be attributed to the fact that direct gaze regression completely ignores the eye geometry which is a valuable prior. Inspired by the recent work in SLAM and scene understanding research where integrating geometry has proved to help deep nets get better at depth estimation~\cite{garg2016unsupervised} and relative pose prediction~\cite{handa2016gvnn, zhou2017unsupervised}, we propose a multi-task deep network that estimates intermediate quantities that geometrically relate to gaze estimation. Our approach has several practical advantages:
\begin{enumerate}
\item Modularity: the estimates from EyeNet work well in conjunction with the classical eye gaze estimation pipeline (see Figure~\ref{fig:pipeline}). We demonstrate that EyeNet predictions can be plugged in to the classical pipeline and vice-versa towards the goal of analyzing gaze estimation performance.   
\item Flexibility: Intermediate results from our multi-stage eye tracking model can drive other vital applications in VR/MR. 3D Cornea center estimates can be used as a proxy for eyeball center of rotation for rendering in waveguide displays; eye segmentation is useful for driving the eyes in avatar rigs for telepresence, emotive expression classification is important for enabling realistic interactions with avatars.
\item Labeling: Ease of ground truthing: collecting large quantities of gaze ground truth for a large number of subjects can be noisy and strenuous. Here we decouple the training of our intermediate predictions (pupil and cornea estimation) from the final 3D gaze vector estimation. The network used for gaze estimation is therefore small and hence we reduce the need for large quantities of ground truth gaze. This idea has also been successful for other visual tasks~\cite{wu2016single}). 
\item Interpretability: Errors in predictions from end-to-end deep networks can be hard to interpret. EyeNet’s estimates at every stage of the eye gaze estimation pipeline helps correlate the effects of each estimate on the final gaze prediction.
\end{enumerate}

Figure~\ref{fig:pipeline} shows our overall network structure. It consists of a feature encoding base network and six task branches for eye parts semantic segmentation, pupil center estimation and glint localization, pupil and glints presence classification, 2D cornea estimation, blink detection and emotive expression classification. We describe each component in detail below.

\subsection{Feature Encoding Layers}
\label{sec:fel}
The feature encoding layers serve as the backbone of the multi-task EyeNet model, and they are shared across every task. We employ ResNet50~\cite{he2016deep}, the standard state-of-art image feature extraction network and a feature pyramid (FPN)~\cite{lin2017feature} to capture information from different scales. The input gray-scale image size is $160\times120$ and we use features from the topmost layer of the encoder (size $20\times15\times256$) as input to the task branches. The full architectural details of EyeNet are provided in the Appendix.

\subsection{Supervised appearance based task branches}
There are three major appearance based tasks in our multi-task learning model. In the following, we discuss their formal definition, sub-network structure and associated training losses. 

\subsubsection{Eye Parts Segmentation}  
\label{sec:eye_parts_segmentation}
Eye parts segmentation is defined as the task of assigning every pixel in the input image a class label from one of Background, Sclera, Iris and Pupil. For this task, we take the last layer feature map from the encoder network (see Section~\ref{sec:fel}) and up-sample it using  deconvolutional layers to the same resolution as the input image, similar to~\cite{badrinarayanan2017segnet,ronneberger2015u}. The resulting four channel output is converted to class probabilities using a softmax layer for each pixel independently. The loss is then a conventional cross-entropy loss between the predicted probability distribution and the one-hot labels obtained from manually annotated ground truth. 
Formally, we are minimizing the following loss for a pixel $x,y$ with ground truth class c and predicted probability $p_k(x,y)$ for the $k^{th}$ class;

\begin{equation}
\mathcal{L}(x, y) = -\sum^{4}_{k=1} I_{x,y}[k==c] \log{p_k(x,y)},
\end{equation}
where $I_{x,y}[.]$ is the indicator function. The overall loss is simply the sum of the losses over all pixels in the image.
The segmentation task serves as a bootstrap phase for training the feature encoder layers as it captures rich semantic information of the eye image. By itself, eye parts segmentation can help the initial phase of any classical pipeline in terms of localizing the search for glints (using iris boundary) and to estimate pupil center (using pupil boundary). It can also be useful for rendering eyes of digital avatars.

\subsubsection{Pupil and Glint Localization}
\label{sec:pupil_glints_detect}
The pupil and glint localization branch gives us the pixel location  of the four glints and pupil center, i.e a total of five keypoints. The network decoder layers for these two tasks are similar to the eye parts segmentation branch and predict a set of five dense maps at the output corresponding to the five keypoints.  Each dense map is normalized to sum to unity across all the pixels. A cross-entropy loss is then calculated across all the pixels of each map during training. Once trained, the location of the center of the pupil or a particular glint is the pixel corresponding to maximum probability at the output. Specifically, we are minimizing the following loss for every keypoint (four glints and one pupil center):

\begin{equation}
\mathcal{L}(keypoint) = -\sum^{}_{x,y} I[x,y] \log{p_{x,y}},
\end{equation}
where $I[.]$ is an indicator function that is zero everywhere except for the ground truth keypoint location, $p_{x,y}$ is the predicted probability of the keypoint location and the summation is over all the pixels in the image. 

\subsubsection{Pupil and Glint Presence Classification}
\label{sec:pupil_glints_classify}
In realistic settings, often glints and/or the pupil center can be occluded by the closing of eyelids, nuisance reflections can appear as glints and for some gaze angles glints may not appear on the reflective corneal surface. Therefore it is important to learn to classify robustly the presence or absence of glints and the pupil center. These predictions effectively gate whether a glint should be used for cornea center estimation and similarly for 3D pupil center estimation. 

We take the topmost layer feature map from the EyeNet encoder, use one convolution layer to reduce the number of feature channels, reshape it to a one dimensional array and add one trainable fully-connected layer of size $1500\times10$ to produce a 5$\times$2 sized output. Each pair represents the presence, absence probability for one of the four glints and/or the pupil center. We use a binary cross-entropy loss to learn from human labeled ground truth.

\subsubsection{Blink detection}
\label{sec:blinks}
Detecting blinks is an appearance based task which is useful to drive multi-focal displays and/or digital avatars. Blinks are usually captured across a sequence of images so it is necessary to use temporal information to distinguish blinks from events such as saccades (rapid sideways movement of the eyes). In general, it is hard to accurately locate the blink event in which the eyes are fully closed, particularly at the standard 30fps frame rate. In other cases, it may be important to detect the onset of blinks to reduce latency between detection and application. We, therefore, use a simple definition of blinks as the state of the eye when the upper eyelid covers over 50\% of the entire pupil region. This we found to be useful as a working definition for non-expert human labelers. Given the aforementioned definition of blinks, we find that features from EyeNet trained for tasks such as eye segmentation transfer well to the blink detection task. 

\begin{figure}[h]
\includegraphics[width=0.48\textwidth,height=\textheight,keepaspectratio]{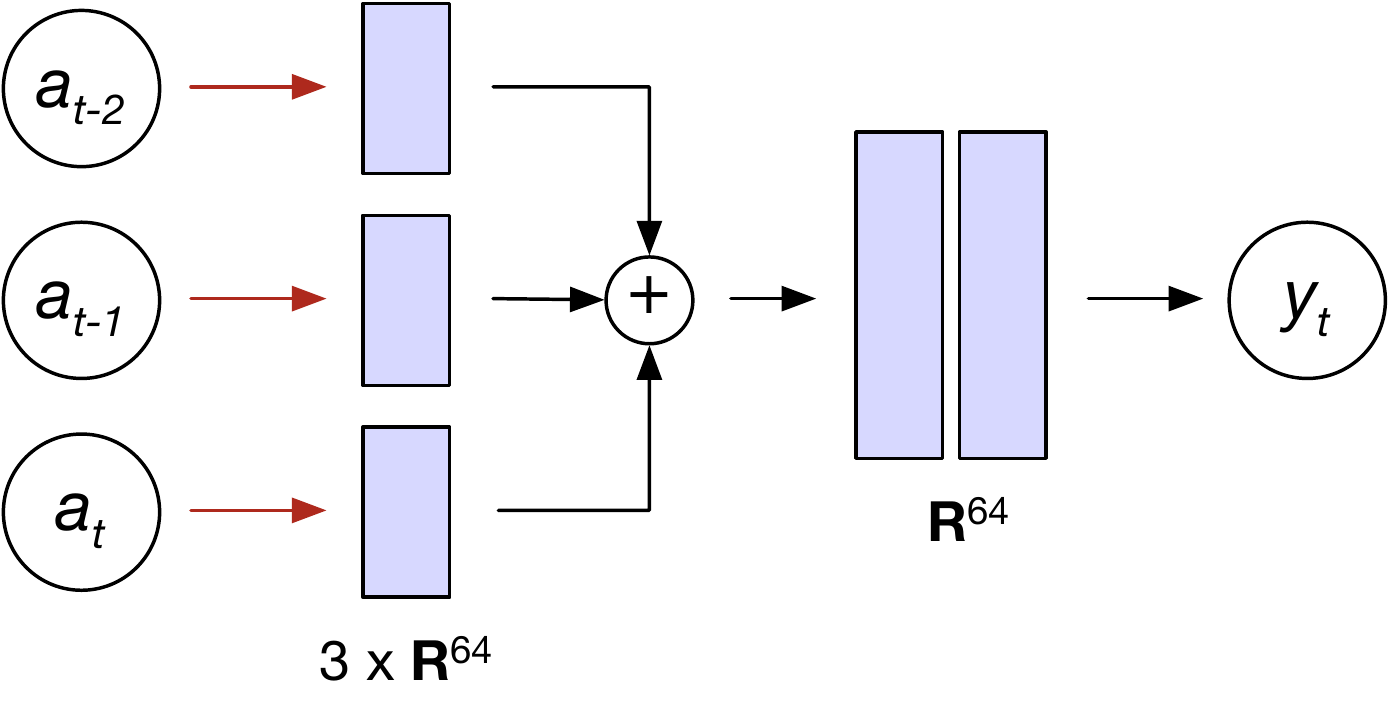}
  \caption{Blink detection at frame $t$. Features from the EyeNet encoder for three consecutive frames ${a}_{t-2},{a}_{t-1},{a}_{t}$ are the inputs. Blue boxes indicate fully connected layers seperated by ReLU, and $\bigoplus$ indicates concatenation.} 
  \label{fig:blink_net}
\end{figure}

We use features from the topmost layer (shared representation) of the pre-trained feature encoding network to train the blink detection branch. As shown in Figure~\ref{fig:blink_net}, the features from three continuous time steps ${a}_{t-2},{a}_{t-1},{a}_{t}$ are fed to a simple 3 layer fully connected network that classifies the current frame (at time $t$) as being a blink or an open eye. We experimented with longer temporal window lengths but this gave diminishing returns in prediction accuracy. RNNs and LSTMs have similar train and test performance, therefore we chose to settle with this simple architecture given the lower compute requirements. 

\subsubsection{Facial Expression Classification}
\label{sec:facial_exp}
The facial expression classification task involves classifying the user's emotive expressions from the input eye images. The task is particularly challenging because we only have the user's eye regions as input rather than the entire face, as used in most emotive facial expressions classification benchmarks~\cite{li2018deep}. To our knowledge, only two previous studies have identified facial expressions from eyes, however, in these studies a view of the eyebrows was available~\cite{lee}~\cite{hickson}. Eyebrows are considered important for emotive expression. The present problem is more challenging, as we aim to estimate expressions without a view of the eyebrows. Here, we consider the following individual emotive facial expressions: happiness, anger, disgust, fear, and surprise. We grouped these into 3 discrete states: positive dimension (happiness), discrimination dimension (anger and disgust), sensitivity dimension (fear and surprise), and a neutral dimension. This classification was motivated by previous work on how humans read complex emotional and mental states from the eye region~\cite{lee}. Due to the fact that it is difficult for even human viewers to detect the subtle differences between individual emotive expressions solely from the eye regions~\cite{lee}, we only report classification results on the 3 dimensions. Like the other task branches, we fixed the main encoder trunk of EyeNet and only trained the facial expressions task branch, consisting of several FC layers, for expression classification. Note that we train this task branch for each subject to produce a personalized model. We found that training a general model for a large population of subjects results in poor accuracy.

\subsection{Model Based Learning for Cornea Center Estimation}
\label{sec:geometry}
The center of the cornea is a geometric quantity in 3D which cannot be observed in a 2D image of an eye. Hence, unlike pupil (center of pupil ellipse) or glint labeling, it is not possible to directly hand label the projected location of the 3D cornea center on the image. We therefore propose a two-step method to train a cornea 2D center prediction branch for EyeNet.
First we use well known geometric constraints and all necessary known/estimated quantities (LED, glints) to generate cornea 2D supervision. Then we train the 2D cornea branch using this model based supervision obtained for each frame. 
 
 Predicting the cornea using the learned network has two main benefits over using geometric constraints during evaluation. First it's more robust because deep networks have a tendency to average out noise during training and standard out-of-network optimization can occasionally yield no convergence. Second it only incurs a small and constant time feed forward compute since the cornea task branch consists of only a few fully connected layers. 

\begin{figure}[h]
  \includegraphics[width=0.5\textwidth,height=\textheight,keepaspectratio]{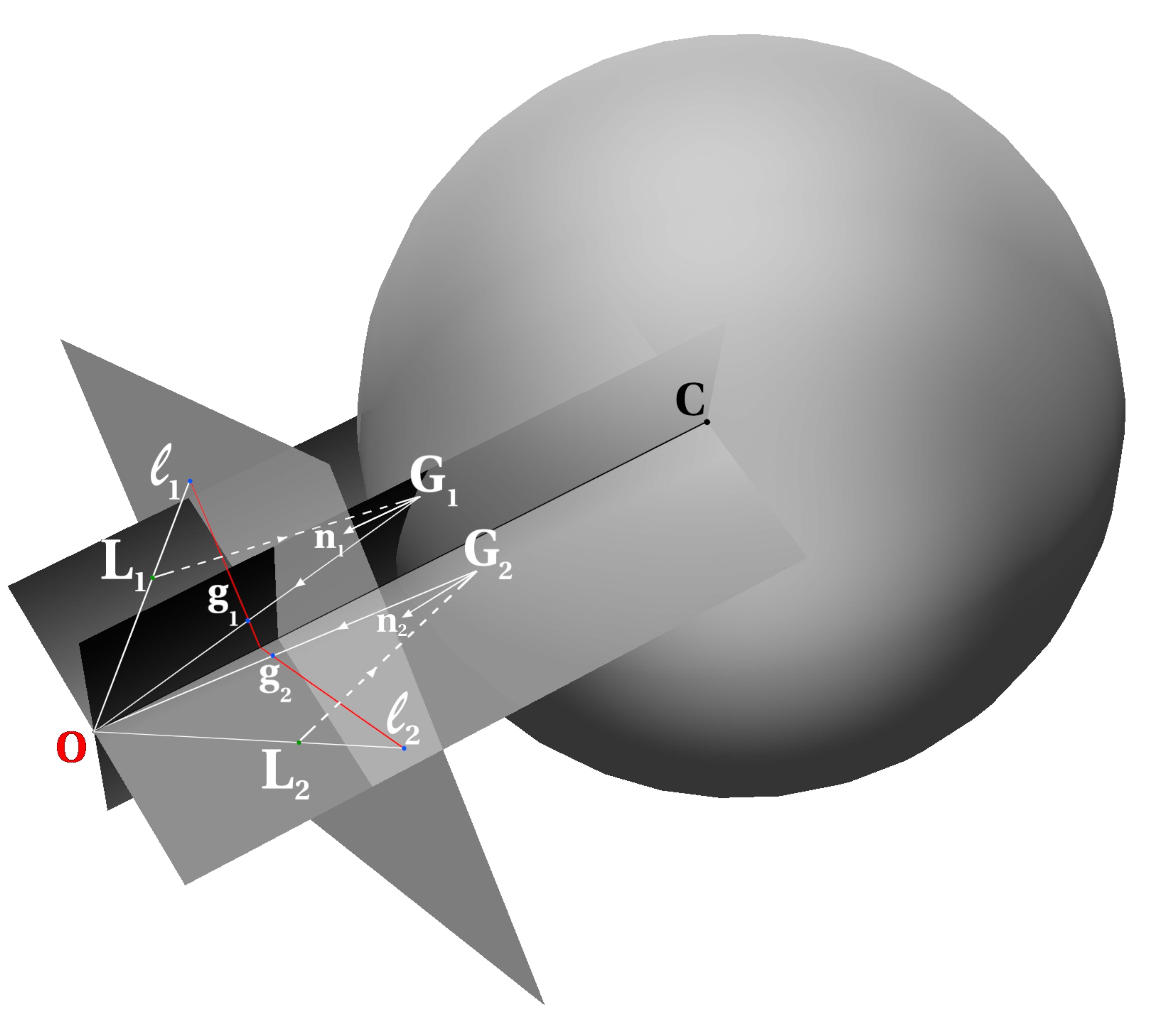}
  \caption{An illustration of the geometric setup used to derive the model based supervision for cornea center training. The corneal sphere along with the camera image plane are shown. According to the law of light reflection, the incident ray $L_{1}G_{1}$, the reflected ray $G_{1}O$ and the normal $n_{1}$ at the point of incidence/reflection are co-planar. The same holds for the other LED. By this constraint, the cornea lies on the ray $OC$ intersecting the two planes.} 
   \label{fig:cornea_geom}
\end{figure}

Figure~\ref{fig:cornea_geom} shows the geometric setup used to derive the model based supervision for training the cornea branch. The corneal sphere, two sample LED locations $L_{1},L_{2}$ in 3D, the camera center $O$ and the image plane are shown. Their incident rays $L_{1}G_{1}$ and $L_{2}G_{2}$ and the corresponding reflected rays $G_{1}O$, $G_{2}O$ can be seen along with the normals $n_{1},n_{2}$ at the points of incidence/reflection. Given the cornea is a reflective surface it follows from the laws of light reflection an incident ray from a point source, e.g. $L_{1}G_{1}$, reflected ray $G_{1}O$ and the normal $n_{1}$ at the point of incidence/reflection on the sphere $G_{1}$ are all co-planar. Equivalently the corneal center $C$, LED $L_{1}$, the projection of the incident point $G_{1}$ onto the image plane denoted $g_{1}$ all lie on the same plane. This holds for other LED point sources too. More details can also be found in~\cite{villanueva2008geometry}. 

In the presence of multiple LED's it is easy to see from Figure~\ref{fig:cornea_geom} that the cornea center lies on the intersecting line $OC$. Line $OC$ lies on all of the planes formed by each pair of LED and corresponding glint normal. This forms a linear system whose solution is the cornea ray. To construct this linear system, we use the cross product of estimated glint normal $OG$ and LED vector $OL$ which provides the corresponding plane equations and then use SVD to solve this linear system. The solution of the linear system is a 3D ray on which the 3D cornea center lies. We intersect this cornea ray with image plane to get the cornea 2D point on an eye image. With this supervision we apply a smooth L1 loss~\cite{girshick2015fast} to train the cornea branch of EyeNet. 

Given the physiology of an individual's eyes is unique, we find it necessary to train the cornea branch to each subject (personalization) in order to produce the most accurate results, sharing the same idea with many recent eye tracking works~\cite{zhang2018training,linden2018appearance}. More details are provided in Section~\ref{sec:cornea_fine_tune}. We now show how to convert the cornea ray prediction from EyeNet to 3D coordinates.

\subsubsection{Cornea Optimization}
\label{sec:cornea_opt}
One of the key elements in the classical pipeline is performing cornea position optimization (both 2D and 3D positions) along with glint detection for each frame starting from a set of blob hypotheses. To do so it uses the geometric model described in Section~\ref{sec:geometry} and a corresponding back projection loss described later in Section~\ref{sec:lifting}. This optimization is performed for each frame independently during test time and directly leads to more accuracy in the final gaze estimate. 

Inspired by this online optimization step, we introduce a variant of EyeNet termed EyeNet-Opt that does a per-frame optimization of the glint and cornea 2D position during test time. The key differences are; 1. unlike the classical optimization which is used to label the glints we rely on the glint labels predicted by EyeNet as our experiments show they are quite robust, 2. we only use simple gradient descent optimization on a loss derived from the eye geometric model in Figure~\ref{fig:cornea_geom}.

\begin{figure}[h]
  \includegraphics[width=0.48\textwidth,height=\textheight,keepaspectratio]{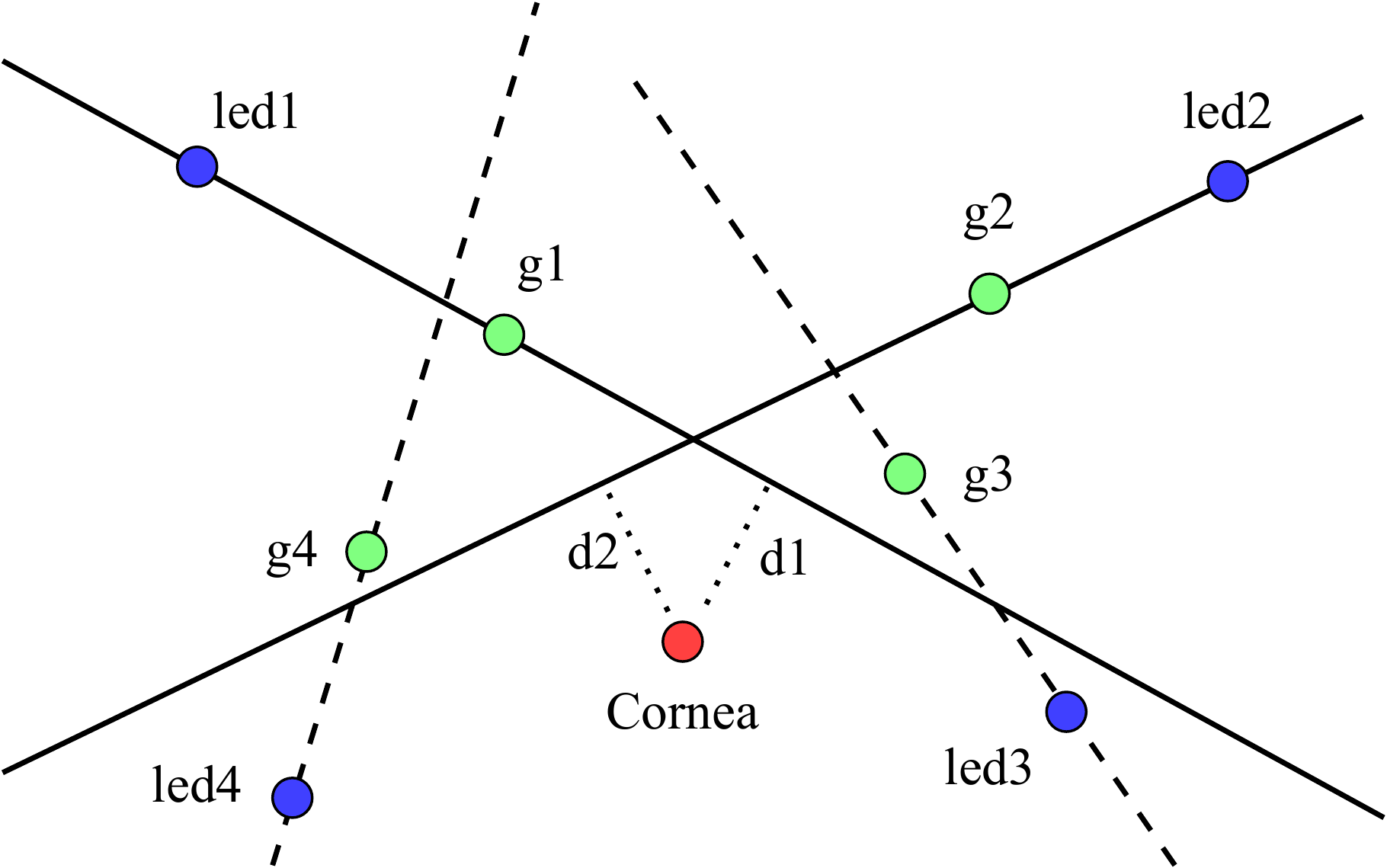}
  \caption{Projections of the LED, cornea 3D location onto the image plane shown alongwith the glint positions. The cornea 2D position should ideally lie on the intersection of all four LED-glint lines. We optimize both cornea 2D and glints positions to minimize the average distance from cornea 2D position to all the glint-led lines.}
  \label{fig:cornea2D}
\end{figure}

From Figure~\ref{fig:cornea_geom} we note that the line joining the projection of an LED onto the image plane $l_{i}$ and its corresponding glint location $g_{i}$ on the image plane intersects the cornea ray $OC$ due to the co-planarity constraints discussed in Section~\ref{sec:geometry}. A pair or more of such lines will intersect at the cornea center projection (cornea 2D) on the 2D image plane. The image plane view of this model based argument is shown in Figure~\ref{fig:cornea2D}. We use this constraint to jointly optimize for the cornea 2D and glint positions on the image.

The loss we employ is minimizing the distance from cornea 2D to all $l_{i},g_{i}$ lines as shown in Figure~\ref{fig:cornea2D}. Here we jointly optimize the cornea 2D location and the glint locations (excluding glint labels) to reach the lowest overall average distance from cornea 2D to all valid $l_{i},g_{i}$ lines. In practice, we perform $100$ steps of gradient descent on each test frame. Our observations are that the initial values of the cornea 2D and glints from EyeNet are robust and lie close to the final value. This allows for confident optimization.

We note in passing here that we also attempted to employ this projection loss to directly train the cornea 2D branch assuming glint detections from EyeNet and/or hand labelled glints. This method unfortunately did not yield satisfactory results.

\subsubsection{Lifting 2D Cornea Center to 3D}
\label{sec:lifting}
\begin{figure}[h]
  \includegraphics[width=0.5\textwidth,height=\textheight,keepaspectratio]{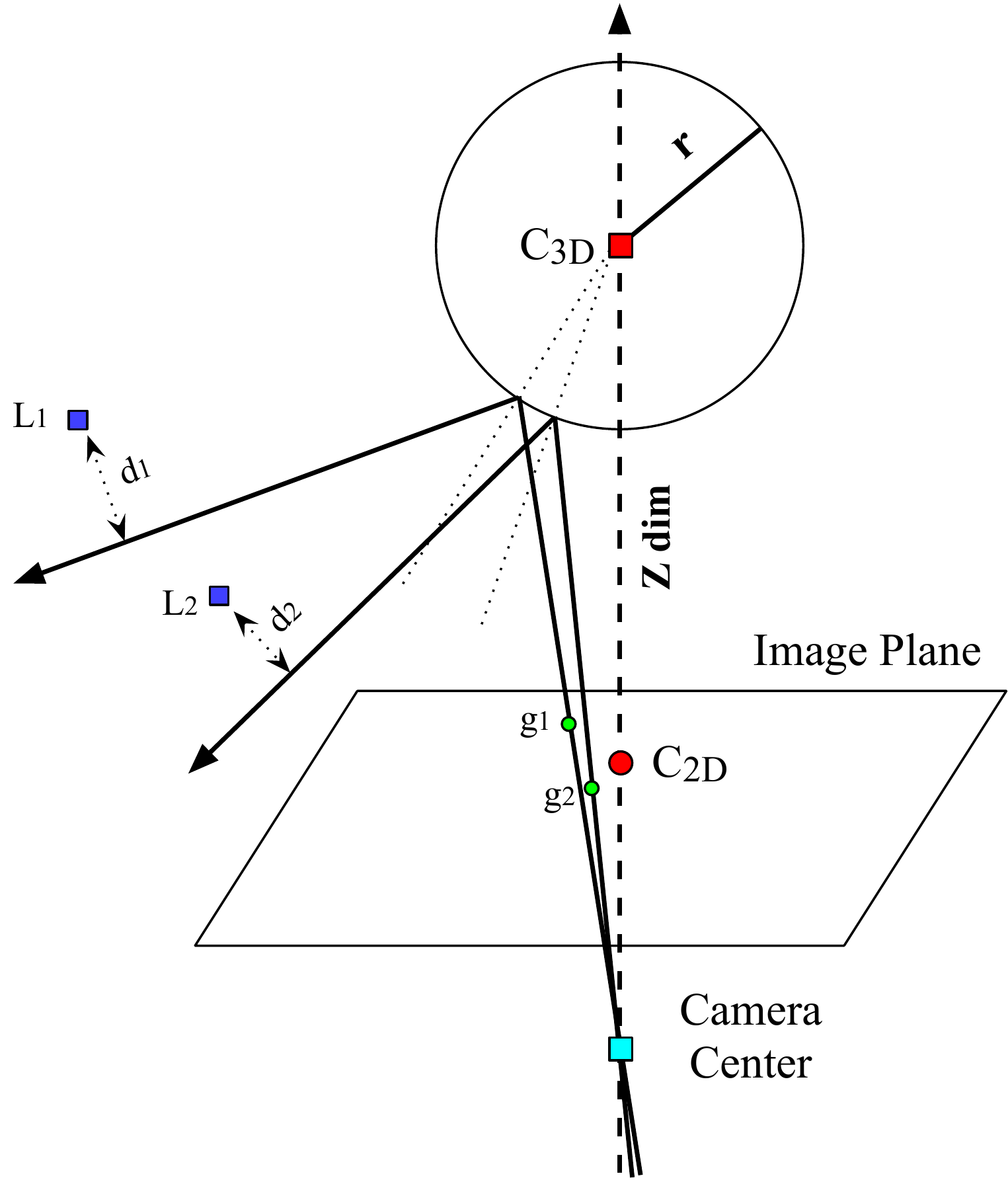}
  \caption{Estimating the cornea 3D position by a line search along the ray connecting the 2D cornea position and the camera center. The optimal position of the 3D cornea, according to the spherical model, is the one which minimizes the distance of the reflected glint rays to the known LED positions. Here the corneal radius is assumed to be 8mm.}
  \label{fig:cornea3D}
\end{figure}

Figure~\ref{fig:cornea3D} depicts the fact that the 3D cornea center lies on a ray passing through the 2D cornea center and the camera center. Therefore, the search for the 3D cornea center is constrained to a single dimension on this cornea ray. At each hypothesized cornea 3D location along this ray, we can compute the ray passing through a 2d glint e.g. $g_{1}$ and its intersection with the corneal sphere (of assumed radius $r$). This ray can be reflected about the normal at the point of intersection $G_{1}$. If the hypothesized corneal 3D location is correct this reflected ray should pass very close to the known LED location $L_{1}$. This idea is converted to a loss function over multiple LEDs and minimized using a discretized search (out-of-network optimization). The details are described below.

We begin by instantiating a sphere of known radius {\lq r\rq} around a hypothesized cornea center in 3D ${C}_{3D}$ (see Eqn \ref{sphere_eqn}) by using the inferred 2D cornea center as well as an initial guess for its {\lq z\rq} dimension: 

\begin{align}
\mathcal{}\left \| x-{C}_{3D} \right \|^{2}_{2} = r^{2}.
\label{sphere_eqn}
\end{align}

As we see from the geometry in Figure~\ref{fig:cornea3D}, a ray from the origin that passes through one of the glints should intersect the cornea sphere if it is located at some nominal camera to eye distance. The direction of such a line is given by the glint unit vector (${\boldsymbol{\hat{\textbf{g}}}}$) in camera centered coordinates. Its line equation is given by, 
\begin{align}
\textbf{y} = {\boldsymbol{\hat{\textbf{g}}}}t,
\label{glint_line}
\end{align}
where 't' is a scalar. If the estimate of the 3D cornea position is correct, this ray can be reflected about a normal at the surface of the cornea which would give us a ray along which the corresponding LED should lie~\cite{spherereflection}. The point of intersection $G_{3D}$ between the glint line and the cornea sphere is given by, 
\begin{align}
{G}_{3D} = {\boldsymbol{\hat{\textbf{g}}}} t^*,  
\label{point_of_intersection}
\end{align}
where $t^*$ is obtained by solving for 't' in Eqn \ref{sphere_eqn} and \ref{glint_line}: 
\newcommand{\mypm}{\mathbin{\smash{%
\raisebox{0.35ex}{%
            $\underset{\raisebox{0.5ex}{$\smash -$}}{\smash+}$%
            }%
        }%
    }%
}
\begin{align}
t^* = {\boldsymbol{\hat{\textbf{g}}}}\cdot{C}_{3D} \mypm \sqrt{({\boldsymbol{\hat{\textbf{g}}}}\cdot{C}_{3D})^2 - ({C}_{3D}\cdot{C}_{3D}-r^2)}.
\label{t_star}
\end{align}
Note that $t^{*}$ must be positive. At the point of intersection, we estimate a normal and a reflected ray as given by the following equations:
\begin{align}
{\boldsymbol{\hat{\textbf{n}}}} = \dfrac{{C}_{3D}-{G}_{3D}}{\left \|{C}_{3D} - {G}_{3D} \right \|},  
\label{normal_eqn}
\end{align}
and
\begin{align}
{\boldsymbol{\hat{\textbf{r}}}} = 2*({\boldsymbol{\hat{\textbf{n}}}}\cdot{\boldsymbol{\hat{\textbf{g}}}}){\boldsymbol{\hat{\textbf{n}}}} - {\boldsymbol{\hat{\textbf{g}}}}.
\label{reflected_ray}
\end{align}
Measuring the distance between the actual LED positions and their respective reflected rays from cornea sphere gives us a loss that we minimize to find the optimal 'z' for the cornea 3D:
\begin{align}
\label{LED_loss}
\mathcal{L}({C}_{3D}) = \dfrac{1}{2}\sum^{4}_{i=1}  {d}_{{L}_{i}}^2({C}_{3D}),
\end{align}
where ${d}_{{L}_{i}}$ is the distance between the reflected ray $\hat{\textbf{r}}$ and the respective LED $L_{3D}$:
\begin{align}
    {d}_{L}({C}_{3D}) =   {\left \| ({G}_{3D}-{L}_{3D}) - \big[({G}_{3D}-{L}_{3D})\cdot{\boldsymbol{\hat{\textbf{r}}}}\big] {\boldsymbol{\hat{\textbf{r}}}}\right \|_{2}}.
\end{align}
Note that in the above equation we have dropped the LED/Glint index for brevity. Also, in practice all the four glints may not be detected due to occlusion or presence of distractors. Therefore the sum in Eqn  \ref{LED_loss} is only over valid glints.
It is important to note that the glint unit vectors would not intersect the cornea sphere if the cornea Z position is too close to or too far from the camera. This implies the loss in Eqn \ref{LED_loss} is not differentiable outside of a small range of camera to eye distances. This makes it difficult to train EyeNet to directly minimize the loss. Therefore, we use a straightforward discretized search along the 1D ray to locate the optimum 3D cornea position. In our setup, as show in Figure~\ref{fig:device} the 'z' distance between the camera and the cornea sphere is usually bounded between 1 and 5 centimeters. We compute the loss described in Eqn \ref{LED_loss} for every ten thousandth of a centimeter between the upper and lower bounds and pick the 'z' value which minimizes the loss.

\subsection{Gaze Estimation}
\label{sec:gaze_estimation}
Following the illustration in Figure~\ref{fig:Setup}, given the estimate of the pupil center on the image plane and the cornea in 3D, we can  connect the camera center and pupil 2D to form the pupil normal, where the pupil center lies. We extend the pupil normal to intersect with the corneal ball to get an approximate pupil center in 3D. By connecting pupil 3D and cornea center 3D, we get an estimate of the optical axis. 

We learn to estimate the gaze direction or equivalently the visual axis by training a gaze mapping network termed as the DeepGazeMapper. This deep network takes as input the optical axis direction and outputs the visual axis direction. Specifically, the visual axis is defined as the line connecting cornea 3D center to the calibrated visual target location.  
In practice, we transform the optical axis and visual axis to the device coordinate system and unit normalize them before training. When the cornea 3D is also transformed to this coordinate system the gaze vector can be drawn through the cornea center and displayed. 

In existing methods, a mapping function in the form of a hand prescribed second order polynomial function is fit to map pupil position to the 3d target position, effectively producing  
the visual axis~\cite{guestrin2006general}. It is also possible to use a second order polynomial function to map the optical axis to the visual axis as is done in our implementation of the classical pipeline. This function is learnt for each subject using the calibration frames. We have shown here that this function can be replaced with a simple fully connected deep network (5 layers and ~30K params) also trained on only the calibration frames, described in experiments section.

\section{Experiments}

\begin{figure*}[h]
  \includegraphics[width=\textwidth,height=\textheight,keepaspectratio]{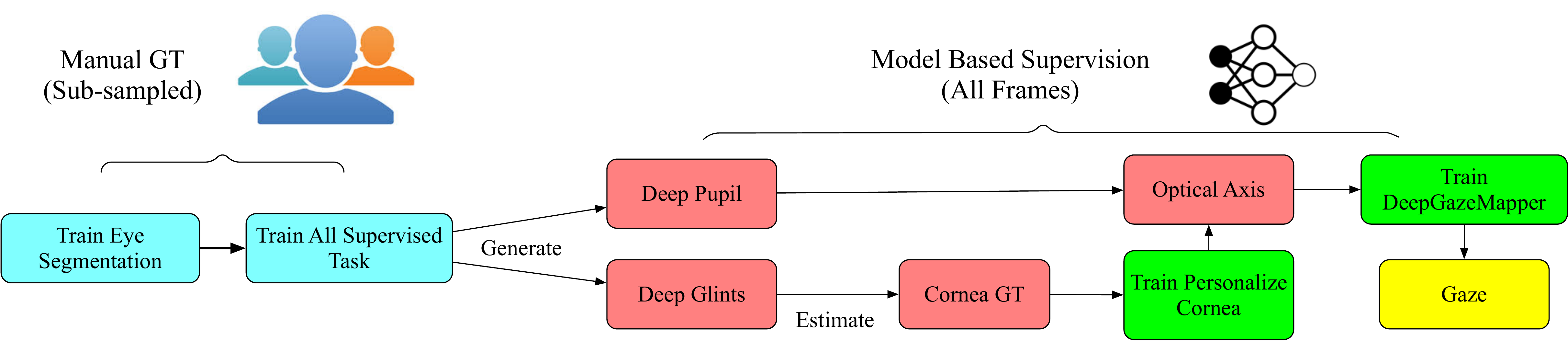}
  \caption{EyeNet training pipeline. Blue boxes indicate steps where all tasks are jointly trained. Green boxes indicate steps where only the specific task branch is fine-tuned and the main encoder is frozen. Red boxes means quantities generated by the network.}
  \label{fig:train}
\end{figure*}

\subsection{Dataset}

The device setup for data collection with an off-axis camera is shown in Figure~\ref{fig:device}. We collected an extensive in-house dataset of $587$ subjects across different genders, demographic groups with different physiology such as skin and eye colors. Table~\ref{table:eye_stats} shows the distribution of subjects across different ethnicity and eye colors. The distribution of subjects is nearly even between both genders.  




\begin{table}[h!]
\centering
\begin{tabular}{|c|c|c|c|c|c|}
\hline
           Brown      & Hazel          & Blue         & Green          & Other    \\ \hline
 63\%          & 16\% & 11\% & 8\%          & 2\%          \\ \hline
\end{tabular}

\vspace{2pt}
\centering
\begin{tabular}{|c|c|c|c|c|}
\hline
           Caucasian      & African descent          & Hispanic         & Other     \\ \hline
 59\%          & 25\% & 12\% &  4\%          \\ \hline
\end{tabular}
\caption{\small{Eye color and ethnicity distribution of subjects.}}
\label{table:eye_stats}
\end{table}

In the data collection phase, each subject faces a virtual 3x3 grid of points at six distinct depths of 0.33m, 0.5m, 1m, 1.5m, 2m, 3m. On a given cue, the subject is asked to focus their gaze on one of these 54 3D points as shown in Figure~\ref{fig:target_diagram}. This provides the ground truth gaze vectors for each frame in the dataset. However, we note that there is diminishing returns in annotating segmentation, glints and pupil centers for every frame at 30 or 60 Hz recordings. We uniformly sample 200 left or right eye image frames in each subject’s recordings to manually annotate segmentation, glint presence or absence, glint 2D and pupil 2D positions. Overall, we have 87,000 annotated images with all types of ground truth in our dataset which is used to train and validate our EyeNet’s results. We split this dataset into 334 training subjects and 93 validation subjects to train the network. We use a test set of 160 subjects with gaze target ground truth (980K images) to demonstrate the network's ability to generalize across different subjects.

\subsection{Labeling}
We presented annotators with 2D eye images and polygon marking, ellipse fitting tools. Each of the eye parts, pupil, iris, sclera and background are segmented by finely marking the vertices of their corresponding polygon. The pupil center is then derived by fitting an ellipse to the pupil polygon vertices and noting its center. 
To annotate the glint locations and labels (association with the corresponding LED) in a particular frame, human labelers use past and future frames to mark the glint blobs and also note the  presence or absence of a glint. Each glint location is then  obtained by fitting an ellipse to the marked glint blobs. The glint labels are assigned by observing their relative positions. 

For blink detection, we collected realistic eye motion and blink data from 65 subjects. The annotators were presented with a temporal sequence of eye images and asked to annotate individual eye frames. The images were labeled as blink if more than 50\% of the pupil is covered. Since it's tricky to accurately annotate 50\% pupil covered cases, the labelers were allowed to label each frame as either blink, open-eye, or unsure. The "unsure" labels were ignored for training and for computing accuracy metrics. A total of 40 subjects were used for training and 25 subjects for test. 

For the facial expression classification, we collected data from 15 subjects. We trained subjects to perform facial action units and labelled these data using the Facial Action Coding system, a widely recognized method for coding individual facial muscle movements~\cite{eckman}. These facial action units can then be mapped onto emotive expressions.

\subsection{Calibration}
\label{Calib}
In the calibration phase for each user we use 9 targets (on the 0.5m plane) out of the 54 to personalize (fine tuning) the 2D cornea branch of EyeNet and the gaze mapping net. The remaining 45 targets are used in the evaluation phase.
\subsection{Training and Testing Procedures}

The complete training takes several steps because our framework receives ground truth from different sources as well as a model based supervision which requires estimates from the trained network itself. Figure~\ref{fig:train} shows the overview of the whole training process. The first step is to train the ResNet~\cite{he2016deep} encoder-decoder network with eye segmentation labels because it provides the richest semantic information and is the most complicated supervised task to train accurately. Secondly, we use human labeled glint, pupil 2D center data \footnote{Obtained by fitting an ellipse to the human labelled pupil boundary.} and eye segmentation data together to jointly train all of these three supervised tasks. We find that initializing with weights trained from eye segmentation results in much more stable training than from random initialization. After this step, we generate glint predictions for all frames in MagicEyes and use these along with known LED locations to generate cornea 2D GT for training the cornea branch as discussed in Section~\ref{sec:geometry}. We emphasize here that the cornea branch is trained with data from the whole training set population. It is further personalized (fine-tuned) at the per subject calibration phase.

We use the predicted cornea (personalized) and pupil 3D centers from the calibration frames to deduce the optical axis. Using the gaze targets ground truth of calibration frames, we train the gaze mapping net to transform the optical axis to the visual axis. During test time, we obtain the predicted cornea and pupil 2D centers from EyeNet. These quantities we lift to 3D (see Section~\ref{sec:lifting}) and get the optical axis, which is then fed into the gaze mapping net to infer the predicted gaze direction.

The blink and facial expression classification tasks are trained on top of intermediate features of the main feature encoding branch. Blink detection is a temporal task, we pass three consecutive eye images and extract their intermediate features. With a set of pre-computed features the blink detection branch is trained separately, while the main feature encoding branch of eyeNet remains frozen. Similar procedure is followed at test time. For facial expression classification, we freeze the main feature trunk and only train the expression classification head using expression data. The expression predictions are produced along with all other tasks during testing time.

\begin{figure*}
  \includegraphics[width=\textwidth,height=\textheight,keepaspectratio]{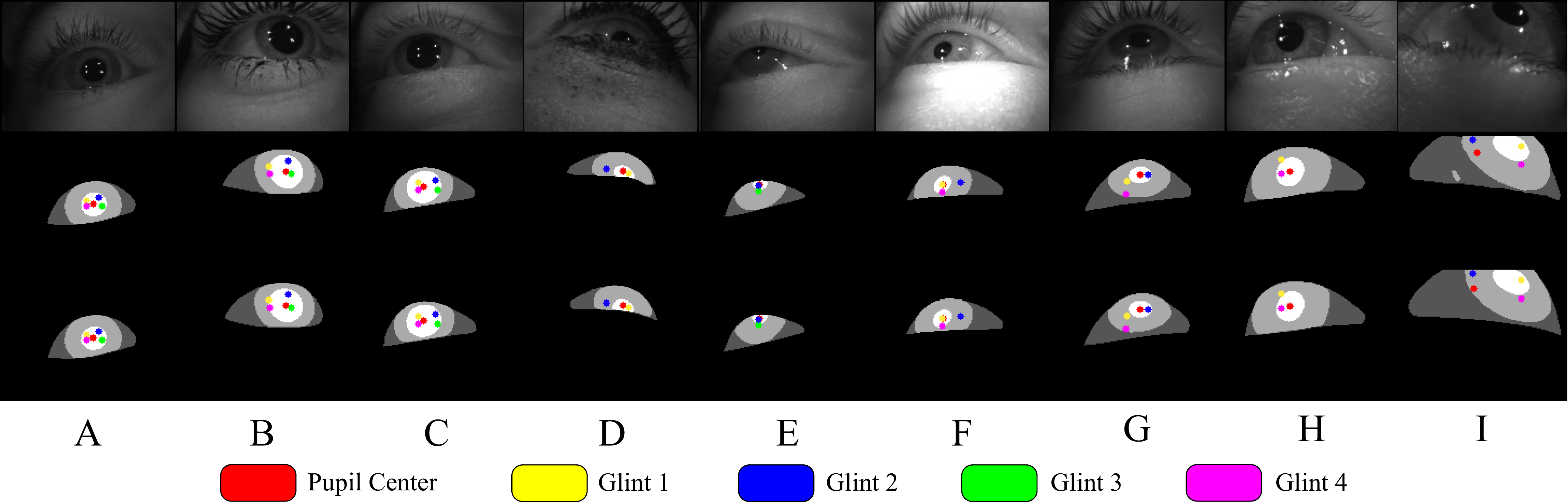}
  \caption{Qualitative Results for EyeNet (Best viewed in color). First row shows sample input eye images. Second row shows the predictions from EyeNet. Third row shows the ground truth obtained from human labelers. Col A,B,C show EyeNet's high quality predictions in frontal facing images. Col D,E demonstrates robustness of the predictions to partial occlusions and small sized pupils. Col F demonstrates the robustness to intensity changes for EyeNet. Col G,H indicate the difficulty faced by human labelers to pinpoint glint locations under significant reflections. However, with sufficient training EyeNet is capable of ignoring nuisance reflections. Col I shows that erroneously labelled pupil center training data can affect EyeNet's predictions.}
  \label{fig:results}
\end{figure*}



\subsubsection{Personalizing Cornea 2D Prediction}
\label{sec:cornea_fine_tune}
Estimating the position of the cornea is a subtle regression task that requires high precision for estimating the 2D location of the cornea (and subsequently its 3D position) so that the visual axis can be estimated accurately. It is a fact that the physiology (e.g radius) of the cornea differs among subjects and hence EyeNet's predictions need to be personalized to each subject for best performance. 

To personalize the cornea 2D branch for each test subject, we  generate predicted glints for all calibration frames of the test subject using the trained EyeNet model. As described in Section~\ref{Calib} we only use a part of the calibration frames of each test subject to fine-tune the cornea 2D branch using the model based supervision (see Section~\ref{sec:geometry}). The remaining non-overlapping set of frames from other calibration targets are used for benchmarking the performance of each personalized EyeNet model. Note also that predictions from this personalized cornea 2D branch provide the initial values for cornea 2D optimization for each frame during test time (see Section~\ref{sec:cornea_opt}).

\subsection{Metrics and Results} \label{sec:results}
We measure the performance of predictions from each branch on the  personalized EyeNet models. These help understand the errors contributed towards the goal of predicting the users gaze in 3D. All the quantitative results are reported by averaging five runs of training and testing.

\subsubsection{Eye Segmentation}
Eye segmentation accuracy is reported using the classic confusion matrix metrics shown below with an averaged trace of 97.29\%. The qualitative results are shown in Figure~\ref{fig:results}.

\begin{table}[h]
\centering
\label{tab:segmentation}
\begin{tabular}{|c|c|c|c|c|}
\hline
GT\textbackslash{}Pred & Pupil          & Iris           & Sclera         & BG             \\ \hline
Pupil                  & \textbf{96.25} & 3.75           & 0.00           & 0.00           \\ \hline
Iris                   & 0.04           & \textbf{99.03} & 0.93           & 0.00           \\ \hline
Sclera                 & 0.00           & 3.27           & \textbf{96.71} & 0.02           \\ \hline
BG                     & 0.01           & 0.72           & 2.09           & \textbf{97.18} \\ \hline
\end{tabular}
\caption{Eye Segmentation confusion matrix in percentage values. The accuracy for all four classes are over \textbf{97.29\%}. EyeNet has very few mis-classification of the pixels to their rightful semantic class.}
\end{table}

The eye segmentation result is very accurate in terms of both quantitative and qualitative evaluations. This is important since the segmentation boundaries are used to generate precise pupil 2D center location training data, particularly for the partially occluded pupil cases, by carefully tuned ellipse fitting procedures. The segmentation predictions are also used by the classical geometric pipeline which we use as a baseline in this work for gaze estimation comparisons.

\subsubsection{Pupil and Glint Detection}
The pupil accuracy is defined as the Euclidean distance between the human labeled (or algorithmically estimated) ground truth pixel 2D center location and the predicted pixel location from EyeNet. For the glints, the order (label) of the glint matters significantly in the cornea estimation phase (glints need to be paired with their corresponding LEDs). We therefore define the metric as the labeled Euclidean error (LEE):
\begin{equation}
LEE = \sum_{i=1}^{4} \lVert G_i - g_i \rVert_{2},
\end{equation}
where the index of the predicted glints $g_i$ is required to be given by the network to match the ground truth glints $G_i$. If the network or classical method labels the glints incorrectly, it would be penalized based on this metric.

The qualitative result of pupil and glint is shown in Figure~\ref{fig:results}. The estimated pupil center is indicated by a red dot, whereas the four glints in order are shown in blue, yellow, pink and green clockwise for left eye (counter-clockwise for right eye).

\begin{figure}[h!]
  \includegraphics[width=0.5\textwidth,height=\textheight,keepaspectratio]{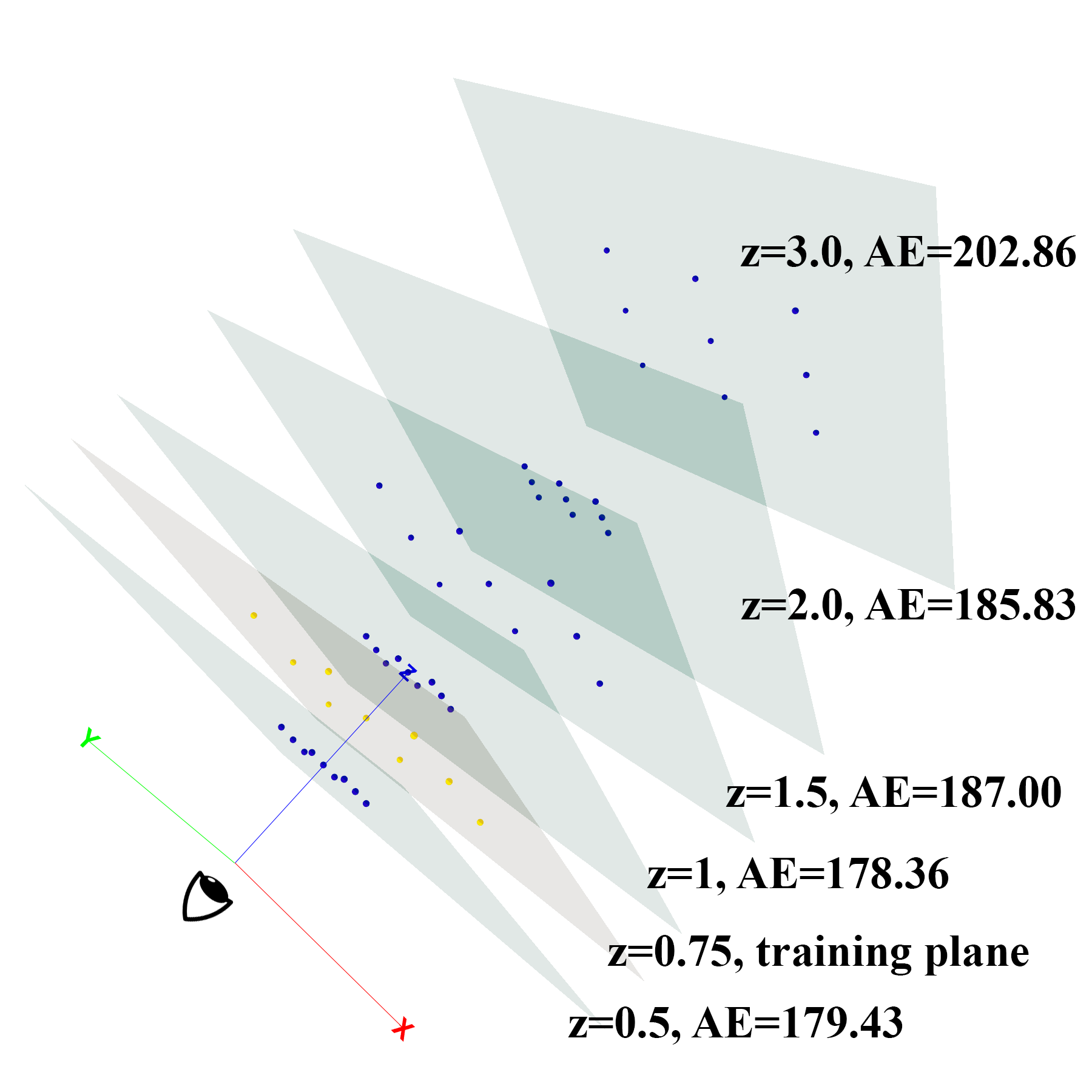}
  \caption{Visualization of gaze targets on each of the 6 planes used in our experiments. The Z value is the distance from the user. The second plane targets are used as calibration targets. As expected the mean angular error increases as depth increases.}
   \label{fig:target_diagram}
\end{figure}

Table~\ref{Pupil_Glint} shows the quantitative result compared to the output of the classical pipeline in pixel errors. When the images are from ideal settings, EyeNet and classical predictions are all precise with close-to zero errors. However, when the images have severe reflections or the users gaze is away from the central targets (see Figure~\ref{fig:target_diagram}), EyeNet is able to first detect the presence or absence of the glints very accurately, and provide robust labeling of the glints, whereas the classical approach suffers from inferior absence indication and mis-labeling of the glints, resulting in much higher error under our labeled Euclidean error metric.

\begin{table}[h]
\centering
\small
\begin{tabular}{|c|c|c|c|c|}
\hline
            & Classical  & EyeNet    & Classical  & EyeNet   \\ 
             & Localization & Localization    & Presence/ & Presence/   \\
             & in pixels & in pixels    & Absence &  Absence \\
            \hline
Pupil       &  0.64     & \textbf{0.46} &  92.81\%                 & \textbf{99.61\%} \\ \hline
Glint 1     &  1.21     & \textbf{0.47} & 90.16\%         & \textbf{96.94\%} \\ \hline
Glint 2     &    1.08           & \textbf{0.39} &            90.84\%       & \textbf{96.32\%} \\ \hline
Glint 3     &    0.84           & \textbf{0.23} &           92.14\%        & \textbf{96.85\%} \\ \hline
Glint 4     &    0.78           & \textbf{0.37} &            91.56\%       & \textbf{96.34\%} \\ \hline
Avg &     0.86          & \textbf{0.38} &  91.72\%                 & \textbf{98.06\%}      \\ \hline
\end{tabular}
\caption{Pupil and glints localization and presence-or-absence (PoA) classification. The first two columns are localization LEE of EyeNet and classical predictions. The final two columns show the PoA classification accuracy. EyeNet provides much more precise PoA classification and this in turn produces more accurate and robust localization.}
\label{Pupil_Glint}
\end{table}
\vspace{0cm}

\begin{figure}[h!]
  \includegraphics[width=0.5\textwidth,height=\textheight,keepaspectratio]{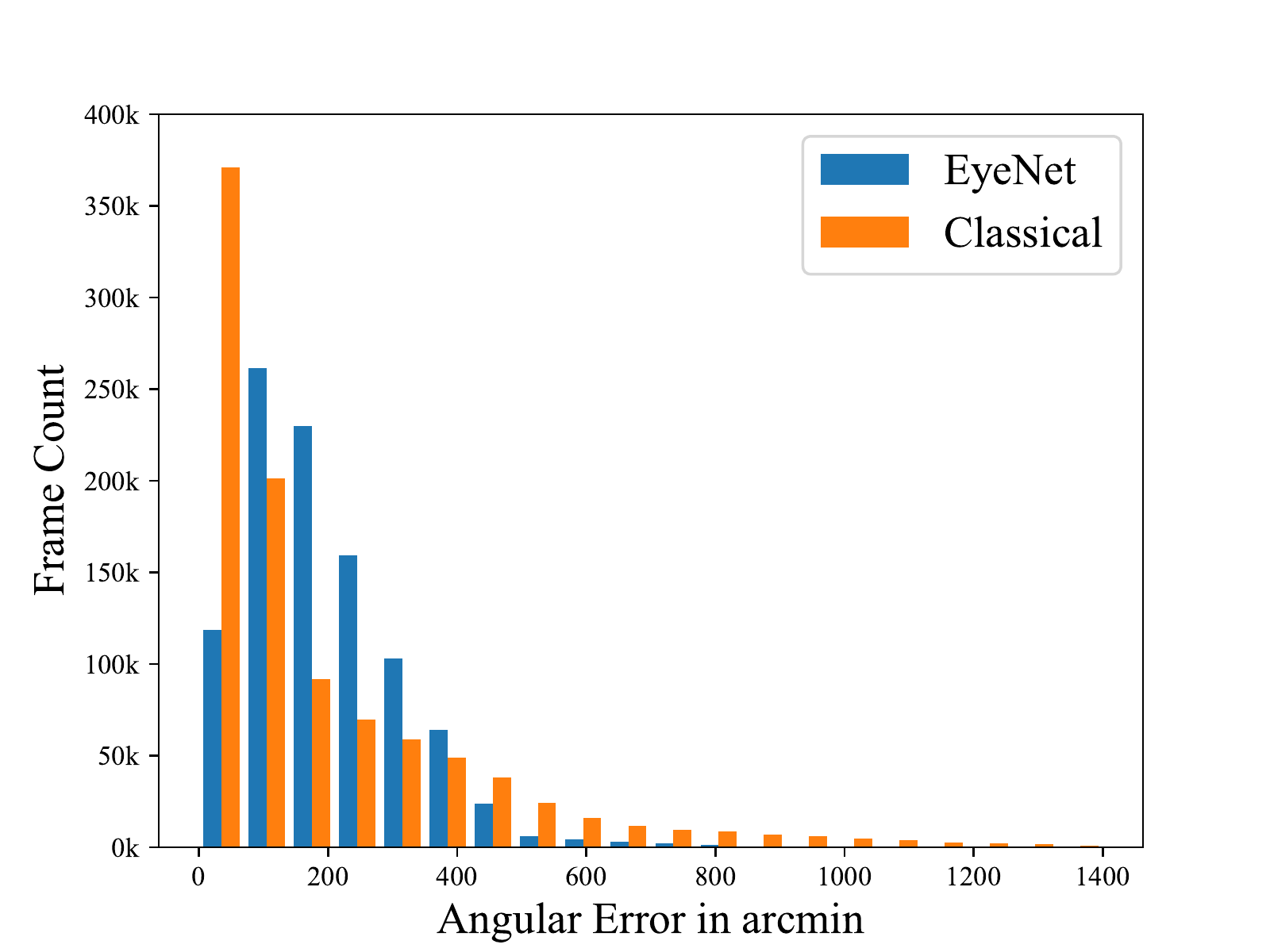}
  \caption{Error histogram for gaze estimation. EyeNet results are more centered near average accuracy without many outliners. Classical approach has more low error frames but is also more spread resulting in potential jitters.} 
  \label{fig:cornea_loss}
\end{figure}

\begin{table*}[h]
\centering
\begin{tabular}{|c|c|c|c|c|c|c|c|c|}
\hline
Model& Description          & Mean AE      & Std AE                & Q1 AE    & Q2 AE       & Q3 AE             \\ \hline
1 &Classical  & 204.98 & 240.17 & \textbf{52.85} & \textbf{101.34} & 290.27  \\ \hline
2 &Classical-DeepGazeMapper & \textbf{183.43} & 153.11 & 65.39 & 131.50 & 264.16 \\ \hline
3 & EyeNet-DeepGazeMapper  & 238.85 & 114.88 & 147.27 & 231.96 & 327.98  \\
\hline
4 &EyeNet-Opt-DeepGazeMapper & 186.41 & \textbf{105.81} & 102.27 & 167.89 & \textbf{258.30}  \\ \hline

5 &EyeNet Glints \& Pupil -SVD Cornea-DeepGazeMapper & 193.51 & 131.53 & 92.56 & 156.51 & 269.05 \\ \hline
\end{tabular}
\caption{Gaze Angular Error (AE) for EyeNet, Classical and mix-and-match models in arcmin units. Note 60 arcmin is 1 degree. Q1, Q2, Q3 correspond to the 25th percentile, median and 75th percentile error respectively. The overall mean and standard deviation of the error is also listed. Here Opt corresponds to cornea 2D optimization using gradient descent with initial values estimated from EyeNet. ClsMapper indicates use of the standard polynomial mapping for gaze estimation. SVD Cornea computes the 3D cornea position given pre-computed glint locations and labels.\\
}
\label{tab:gaze}
\end{table*}

\begin{table*}[]
\centering
\begin{tabular}{|c|c|c|c|c|}
\hline
              & Deep Mean & Classical Mean & Deep Std & Classical Std \\ \hline
Top Left      & 194.18    & 261.05         & 105.63   & 304.42        \\ \hline
Top Middle    & 169.64    & 148.28         & 103.53   & 143.25        \\ \hline
Top Right     & 184.95    & 162.57         & 109.16   & 154.25        \\ \hline
Center left   & 195.15    & 298.44         & 105.78   & 331.17        \\ \hline
Center Middle & 183.57    & 147.15         & 106.18   & 143.11        \\ \hline
Center Right  & 193.35    & 161.74         & 108.62   & 151.77        \\ \hline
Bottom Left   & 205.55    & 300.94         & 105.56   & 323.00        \\ \hline
Bottom Middle & 179.15    & 154.19         & 100.55   & 146.39        \\ \hline
Bottom Right  & 181.35    & 166.15         & 103.47   & 161.19        \\ \hline
\end{tabular}
\caption{Gaze error for each of the 9 target direction aggregated over all the target planes. It is clear that EyeNet predictions are have smaller and consistent standard deviation across the field of view. This is due to the robustness of EyeNet's estimates and the use of the DeepGazeMapper.}
\label{tab:by_target}
\end{table*}

\subsubsection{Cornea Center Estimation}
It is hard to obtain ground truth for the cornea center. Therefore we compare the Euclidean distance of the EyeNet predicted cornea center to the cornea center obtained by the classical pipeline. On an  average, the direct EyeNet cornea predictions are \textbf{1.77mm} away from the classical pipeline predictions with a standard deviation of \textbf{2.97mm}. This distance is reduced post cornea optimization to \textbf{0.99mm} with a standard deviation of \textbf{2.10mm}, a clear improvement that also boosts the final gaze accuracy shown in Table~\ref{tab:gaze}. The high standard deviation is due to a few high error outliers, which are averaged out by the deep gaze mapping network during the gaze estimation phase. 

It is important to note here that the classical cornea center estimate may not be a good approximation to the true center of rotation of the eyeball from which the gaze is referenced. Indeed, getting closer to the classical estimate improves the final gaze estimate but more work needs to be done to estimate a 'fixed point' for gaze computation which is stable over the desired field of view of the gaze driven application.


\subsubsection{Gaze Estimation}
The overall gaze estimation metric is defined as the angular error between the true gaze vector and the estimated gaze vector in arcmin units. Figure~\ref{fig:target_diagram} shows the targets used in our experiments as coming from various depths. Note that there are only 18 unique directions although there are 54 targets. This setup was chosen to study the gaze errors in the same direction over increasing depths. Table~\ref{tab:gaze} compares the performance of EyeNet to the ubiquitous classical baseline and other mix-and-match variants.

From the results in Table~\ref{tab:gaze} we can see that the Classical pipeline (model 1) has the lowest 25th percentile and median error. However, both the overall mean and standard deviation (std) is high. The use of the DeepGazeMapper significantly lowers the error variance, including that for the classical model (model 2), with only a small impact to accuracy. This shows the DeepGazeMapper by itself can provide significant improvements to stability over the standard polynomial mapping. In most applications which use eye tracking in HMDs it is generally preferrable to avoid jittery tracking while compromising a little on the accuracy. 

EyeNet (model 3) estimates have low std but overall higher error than classical. This implies that joint end-to-end learning for  estimates such as the cornea 2D, glints, pupil etc. can be used for applications which do not have severe accuracy demands. This result however is still a significant improvement over recent attempts to directly regress gaze by end-to-end learning~\cite{shrivastava2017learning}. When the accuracy demands are higher, it is prudent to use these estimates as starting values for simple gradient descent optimization for the 3D cornea position (see Section~\ref{sec:cornea_opt}). As seen from the results for model 4 this improves the accuracy significantly while also delivering improved robustness. 

If the main goal is robust and accurate gaze estimation in a narrow field of view (typically corresponding to data which make up the Q1, Q2 percentiles) then using EyeNet pupil and glint estimates to derive the cornea position out-of-network using SVD lowers the gaze error as in model 5. However the demands on eye tracking will be over a larger field of view corresponding to newer displays on HMDs.

In Table~\ref{tab:by_target} we report the angular error per direction (averaged over all depths) for models 1 and 4. It is clear that model 4 standard deviation is significantly better and similar in all directions. This can primarily be attributed to robust glint, cornea 2D estimates along with the use of the DeepGazeMapper. 

Figure~\ref{fig:target_diagram} shows the gaze error for model 4 over the different planes. It is clear that the error increases with depth but there is approximately 20 arcmin increases between the targets at 0.5m and 3m. This once again indicates the robustness of EyeNet.

The histogram of gaze angular errors for both model 1 and 4 are displayed in Figure~\ref{fig:cornea_loss}. Here we see the EyeNet error distribution has a shorter tail than the classical pipeline distribution. The classical pipeline has better accuracy, primarily due to iterative optimization but poorer robustness as indicated by the very long tail in the histogram.

Based on all these experiments and result analysis, we find that accurately estimating cornea position from eye images remains the most sensitive and challenging part in the whole pipeline. During the training runs we find that outliers in the training data have a significant impact on cornea 2D estimation for example. With more carefully designed gaze data collection protocols we believe its possible to further reduce the error in EyeNet estimates towards the goal of gaze estimation.

\subsubsection{Blink Detection}
The blink detection branch takes in EyeNet encoder features for 3 consecutive frames which help to noticeably reduce false positive blink predictions that occur when using a single eye image. Saccades and other partially closed eye events are better distinguished in our temporal blink detection network. On a test set of 25 subjects, the blink detection branch of EyeNet has a false positive rate of 1.24\% and a false negative rate of 4.01\%, indicating the success of sharing features across multiple tasks. In comparison~\cite{LSLWB18} reports a false positive rate of 8.3\% and a false negative rate of 16.7\% using the publicly available Pupil capture~\cite{pupilcapture} software. 

\subsubsection{Facial Expression Classification}
\begin{table}[]
\centering
\small
\begin{tabular}{|c|c|c|c|c|}
\hline
GT/Pred        & Neutral        & Happy          & Discri. & Sensitivity    \\ \hline
Neutral        & \textbf{94.92} & 0.71              & 2.46           & 1.90           \\ \hline
Happy          & 5.85           & \textbf{93.74} & 0.17           & 0.25              \\ \hline
Discrimination & 6.57          & 0.52           & \textbf{92.64} & 0.28              \\ \hline
Sensitivity    & 9.78          & 0.23              & 0.30           & \textbf{89.69} \\ \hline
\end{tabular}
\caption{Emotive expression classification confusion matrix in percentage. The mis-classification to neutral class is mainly due to data and label noise.}
\label{emotion_classification}
\end{table}

We show accurate emotive facial expression classification into 3 discrete dimensions (Happy, Discrimination, Sensitivity), and Neutral. We report the classification in a standard confusion matrix below and an overall accuracy of 92.75\%.

The expression classification is reasonably accurate given that we are predicting it solely using the eye region. There is also a certain degree of data noise as it is difficult for some subjects to produce eye expressions that differ between classes. 

\begin{figure}[h]
 \begin{center}
 \begin{tabular}{ccc}
  \includegraphics[width=0.14\textwidth]{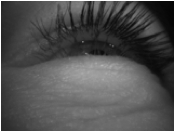} & \includegraphics[width=0.14\textwidth]{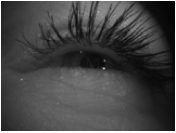} &
  \includegraphics[width=0.14\textwidth]{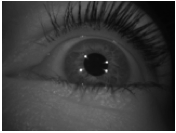} \\
  happy & discrimination & sensitivity
  \end{tabular}
  \end{center}
  \caption{Sample eye images of various expression dimensions.} 
  \label{fig:eye_expressions}
\end{figure}

 The ability to recognize emotive expressions with the eyebrows obscured is important because eyebrows are generally considered critical for emotion expression~\cite{lee}. The positive dimension captures smiling (happiness), the discrimination dimension captures the narrow-eyed state of suspicion (anger and disgust), and the sensitivity dimension captures the wide-eyed state of awe (fear and surprise). These categories are consistent with previous classifications where eyebrow data was available~\cite{lee}. In Figure~\ref{fig:eye_expressions}, we showcase sample eye images of positive, discrimination, and sensitivity dimensions.  

\section{Computational Efficiency}
We benchmark our model running time on a Linux machine with dual Intel E5-2660 CPU and one Nvidia 1080 Ti GPU. We use Pytorch along with  CUDA 9 and CuDNN 7 as the software frameworks. The average running time is \textbf{12ms} or \textbf{83fps} for all EyeNet tasks inference calculated over 1000 iterations. The choice of ResNet50\cite{he2016deep} over ResNet101 or ResNet152 and an input size of 160x120 instead of higher resolution is to be able to deploy the model to AR/MR hardwares for application in the near future. Please refer to the appendix for the full network architecture.

\section{Conclusion and Future Work}

In this work, we presented EyeNet, the first single deep neural network tackling  multiple tasks related to eye gaze estimation and semantic user understanding. We showed that it is indeed possible to learn a shared representation to produce robust estimates for both semantic and geometric tasks simultaneously. This results in competitive gaze accuracy for off-axis gaze estimation with a much improved robustness. Thus, it is not necessary to hand engineer modules for geometric eye gaze estimation or blink, and facial expression detection. The muti-task learning approach utilizes the inductive bias among the relevant tasks to improve performance. Als, the ensuing advantage is the simplicity and maintanability of the solution. Finally, an additional important benefit is that with more training data, EyeNet estimates can be improved further without much expert supervision. 

Recent works attempting to perform end-to-end learning for gaze estimation have resulted in robust but inaccurate estimates. In contrast, we chose to employ domain knowledge in the form of eye geometry for deriving model based supervision. This helps achieve a much higher accuracy in gaze estimation through high quality cornea center prediction. While EyeNet is the first attempt to unify the solution to several problems related to eye gaze estimation and semantics, there are still some limitations such as the need to do out-of-network optimization for achieving higher accuracy. Our next steps are to learn this optimization itself along with the DeepGazeMapper. This presents an opportunity to jointly learn to estimate a stable center of rotation of the eye ball along with the gaze for each frame. Other important considerations for us in the near future is reducing noise during data capture, exploiting synthetic data, incorporating more personalized anatomical and temporal constraints during training. 


\section*{Acknowledgements}
Special thanks to our colleagues Dan Farmer, Brad Stuart, Sergey Prokushkin and Jean-Yves Bouguet for providing us with their valuable insights on classical geometry based eye gaze estimation. Thanks also to Alex Kim for his help in rendering figures explaining eye geometry.

{\small
\bibliographystyle{ieee}
\bibliography{main}
}

\appendix
\section{Appendix}
\subsection{Encoder Architecture}
We emplot a ResNet50 with a feature pyramid network (FPN) as the encoder architecture of EyeNet~\cite{he2016deep},\cite{lin2017feature}. This part can be replaced with newer architectures for potential improvements. We select the pyramid layer with stride of 8 pixels with respect to the input image. Thus, the topmost layer encoder features have a size of 20x15x256. This shared feature is input to the task branches.

Note also that the original image capture size in our dataset is 640x480. However, we downsample it to \textbf{160x120} for computational efficiency as the input to the encoder.

\subsection{Decoder Architectures}
We tabulate each task decoder (branch) in the following. All the components are standard. Note that the Cornea 2D center estimation and facial expression classification share the same structure but they are trained separatedly as two branches. Also see Section~\ref{sec:blinks} for the blink task branch architecture.

\begin{table*}[!htbp] 
\centering
\begin{tabular}{|c|c|c|c|c|}
\hline
Layer Name     & Parent Layer   & Layer Params       & Input Size & Output Size \\ \hline
DeConv1        & Encoding Layer & 14x4x128(stride 2) & 20x15x256  & 40x30x128   \\ \hline
ResidualBlock1 & DeConv1        & 128                & 40x30x128  & 40x30x128   \\ \hline
ResidualBlock2 & ResidualBlock1 & 128                & 40x30x128  & 40x30x128   \\ \hline
DeConv2        & ResidualBlock2 & 4x4x64(stride 2)   & 40x30x128  & 80x60x64    \\ \hline
Conv1          & DeConv2        & 3x3x32             & 80x60x64   & 80x60x32    \\ \hline
BN1+ReLU       & Conv1          & NA                 & 80x60x32   & 80x60x32    \\ \hline
Conv2          & BN1            & 3x3x16             & 80x60x32   & 80x60x16    \\ \hline
Conv3          & Conv2          & 3x3x8              & 80x60x16   & 80x60x8     \\ \hline
DeConv3        & Conv3          & 4x4x4(stride 2)    & 80x60x8    & 160x120x4   \\ \hline
\end{tabular}
\caption{The architecture of the eye parts segmentation branch of EyeNet. See Section~\ref{sec:eye_parts_segmentation} for more details of this task.}
\end{table*}

\begin{table*}[!htbp]
\centering
\begin{tabular}{|c|c|c|c|c|}
\hline
\textbf{Layer Name} & \textbf{Parent Layer} & \textbf{Layer Params} & \textbf{Input Size} & \textbf{Output Size} \\ \hline
Conv1\_Loc          & DeConv2               & 3x3x32                & 80x60x64            & 80x60x32             \\ \hline
BN\_Loc +ReLU       & Conv1\_Loc            & NA                    & 80x60x32            & 80x60x32             \\ \hline
Conv2\_Loc          & BN\_Loc               & 3x3x20                & 80x60x32            & 80x60x20             \\ \hline
DeConv\_Loc         & Conv2\_Loc            & 4x4x5                 & 80x60x20            & 160x120x5            \\ \hline
\end{tabular}
\caption{The architecture of the pupil and glint detection branch of EyeNet. See Section~\ref{sec:pupil_glints_detect} for more details of this task.}
\end{table*}

\begin{table*}[!htbp]
\centering
\begin{tabular}{|c|c|c|c|c|}
\hline
\textbf{Layer Name} & \textbf{Parent Layer} & \textbf{Layer Params} & \textbf{Input Size} & \textbf{Output Size} \\ \hline
Conv1\_Cls          & Encoding              & 3x3x5                 & 20x15x256           & 20x15x5              \\ \hline
Flatten\_Cls        & Conv1\_Cls            & NA                    & 20x15x5             & 1500                 \\ \hline
FC\_Cls             & Flatten\_Cls          & 1500x10               & 1500                & 10                   \\ \hline
\end{tabular}
\caption{The architecture of the pupil and glint presence/absence classification branch of EyeNet. See Section~\ref{sec:pupil_glints_classify} for more details of this task.}
\end{table*}

\begin{table*}[!htbp]
\centering
\begin{tabular}{|c|c|c|c|c|}
\hline
\textbf{Layer Name} & \textbf{Parent Layer} & \textbf{Layer Params} & \textbf{Input Size} & \textbf{Output Size} \\ \hline
Conv1\_Cor          & Encoding              & 3x3x64                & 20x15x256           & 20x15x64             \\ \hline
BN1\_Cor+ReLU       & Conv1\_Cor            & NA                    & 20x15x64            & 20x15x64             \\ \hline
Conv2\_Cor          & BN1\_Cor              & 3x3x16                & 20x15x64            & 20x15x16             \\ \hline
BN2\_Cor+ReLU       & Conv2\_Cor            & NA                    & 20x15x16            & 20x15x16             \\ \hline
Flatten             & BN2\_Cor              & NA                    & 20x15x16            & 4800                 \\ \hline
FC1\_Cor            & Flatten               & 4800x64               & 4800                & 64                   \\ \hline
FC2\_Cor            & FC1\_Cor              & 64x16                 & 64                  & 16                   \\ \hline
FC\_Out\_Cor        & FC2\_Cor              & 16x2                  & 16                  & 2                    \\ \hline
FC\_Out\_Expre      & FC2\_Cor              & 16x4                  & 16                  & 4                    \\ \hline
\end{tabular}
\caption{Cornea 2D center estimation and the facial expression classification share this architecture but are trained separately as two branches. See Section~\ref{sec:geometry} for more details of this task.}
\end{table*}

\end{document}